\documentclass[10pt,journal,cspaper,compsoc]{IEEEtran}
\usepackage{graphicx}
\usepackage{comment}
\usepackage{amsmath,amssymb} 
\usepackage{color}

\usepackage{multirow}
\usepackage{paralist}

\usepackage{times}
\usepackage{epsfig}
\usepackage{algorithm}
\usepackage{algorithmic}
\usepackage[font=small]{caption}

\usepackage{bbm}
\usepackage{wrapfig}

\newcommand{\posnegns}{{\em pos}/{\em neg}}
\newcommand{\posneg}{\posnegns~}

\newcommand{\bpi}{\boldsymbol{\pi}}
\newcommand{\real}{\mathbb{R}}
\newcommand{\indicator}{\mathbbm{1}}

\newcommand{\zcy}[1]{{#1}}
\newcommand{\note}[1]{\textcolor{cyan}{[ZCY: #1]}}

\newcommand{\abc}[1]{{#1}}
\newcommand{\abcn}[1]{{#1}}

\newcommand{\CUT}[1]{}

\begin{document}
	
	\title{Density-based Object Detection in Crowded Scenes} 
	
	\author{Chenyang~Zhao, Jia~Wan, and Antoni~B.~Chan 
		\IEEEcompsocitemizethanks{\IEEEcompsocthanksitem Chenyang~Zhao, Jia Wan, and Antoni~B. Chan (corresponding author)  are with the Department of Computer Science, City University of Hong Kong.\protect\\
			E-mail: zhaocy2333@gmail.com, jiawan1998@gmail.com, abchan@cityu.edu.hk.
		}
		\thanks{}}
	
	\markboth{Journal of \LaTeX\ Class Files,~Vol.~X, No.~X, XXX~XXXX}%
	{Shell \MakeLowercase{\textit{et al.}}: Bare Demo of IEEEtran.cls for Computer Society Journals}
	
\IEEEcompsoctitleabstractindextext{


\begin{abstract}

Compared with the generic scenes, crowded scenes contain highly-overlapped instances, which result in: 1)  more ambiguous anchors  during training of object detectors, and 2) more predictions are likely to be mistakenly suppressed in post-processing during inference. 
To address these problems, we propose two new strategies, density-guided anchors (DGA) and density-guided NMS (DG-NMS), which uses object density maps to jointly compute optimal anchor assignments and reweighing, as well as an adaptive NMS.
Concretely, based on an unbalanced optimal transport (UOT) problem, the density owned by each ground-truth object is transported to each anchor position at a minimal transport cost. And density on anchors comprises an instance-specific density distribution, from which DGA decodes the optimal anchor assignment and re-weighting strategy. Meanwhile, DG-NMS utilizes the predicted density map to adaptively adjust the NMS threshold to reduce mistaken suppressions. 
In the UOT, a novel overlap-aware transport cost is specifically designed for ambiguous anchors caused by overlapped neighboring objects.
Extensive experiments on the challenging CrowdHuman \cite{shao2018crowdhuman} dataset with Citypersons \cite{zhang2017citypersons} dataset demonstrate that our proposed density-guided detector is effective and robust to crowdedness. The code and pre-trained models will be made available later.
\end{abstract}

}

\maketitle

\IEEEdisplaynotcompsoctitleabstractindextext
\IEEEpeerreviewmaketitle

\section{Introduction}

The main stream frameworks widely used in object detection systems generate predictions based on anchors (i.e., anchor points for anchor-free detectors and pre-defined anchor boxes for anchor-based detectors), for both one-stage \cite{2016ssd,2016yolo,2017focal,2019fcos} and two-stage \cite{2015fastrcnn,2017maskrcnn,cai2018cascade,lin2017fpn, dai2017deformable, dai2016r} CNN-based methods. The paradigm typically generates bounding box proposals in a dense detection manner by regressing offsets for each anchor. Therefore,  anchor assignment and re-weighting, which select positive samples, define ground-truth (GT) objects and offer weights for anchors, are necessary during the training of a detector.
Moreover, methods such as non-maximum suppression (NMS) are usually performed to remove duplicated predictions in post-processing.

Many research efforts have achieved  progress on general object detection by improving the anchor assignment strategy \cite{zhang2019freeanchor,2020atss,ke2020mal,kim2020paa,zhu2020autoassign,ge2021ota, wang2021end} and NMS \cite{bodla2017soft,hosang2017learning,he2019bounding}. However, crowded object detection is still challenging in practice due to heavily overlapped instances; compared with the general situation,  crowded overlapping objects results in  more ambiguous anchors during training and the predictions are very likely to be mistakenly suppressed by NMS in post-processing. To effectively detect and preserve high-overlapped positive samples, the current crowded detection methods mainly focus on the latter issue by designing occlusion-aware NMS \cite{liu2019adaptive, hosang2016convnet, zhang2019double,huang2020nms, gahlert2020visibility, chu2020detection}, modifying the loss function for tighter boxes \cite{zhang2018occlusion, wang2018repulsion}, or aiding NMS with visible region information \cite{zhang2019double, huang2020nms, gahlert2020visibility}.   
 
The ambiguous anchor assignment problem caused by overlapping objects is rarely specifically considered in the training of crowded object detectors. Most previous methods follow the anchor assignment strategies of the general object detection baselines.  The classic strategy for generating positive and negative (\emph{pos}/\emph{neg}) anchors is through hand-crafted rules based on human prior knowledge: 1) Anchor-based methods like RetinaNet \cite{2017focal}, region proposal network (RPN) in Faster RCNN \cite{2015fasterrcnn} and FPN \cite{lin2017fpn} assign \posneg using a pre-defined threshold on the IoUs between anchors and GT boxes; 2) Anchor-free methods like FCOS \cite{2019fcos} generally choose a fixed portion of the center area of the GT bounding box as positive spatial positions.
Although these assignment strategies are intuitive and popularly adopted, they ignore the actual content variation across scenarios, which results in sub-optimal assignments.  

\begin{figure}[t]
	\begin{center}
		\includegraphics[width=0.45\textwidth]{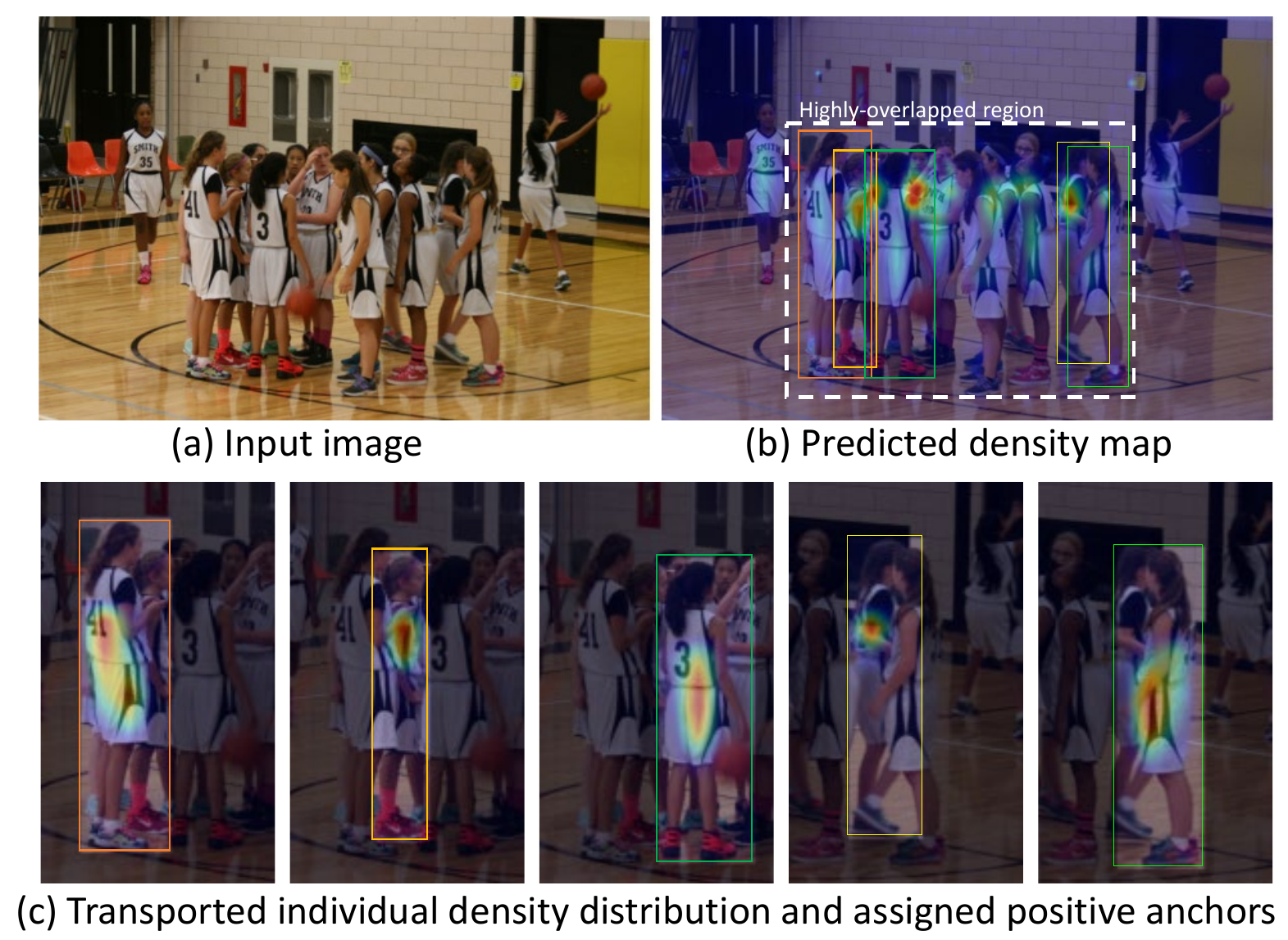}
	\end{center}

	\caption{An example of predicted density map and assigned individual density map for  instances in highly-overlapped region. 
	}
	\label{fig:figure1}

\end{figure}

Recent works aim to address the limitation of hand-crafted anchor assignment by proposing dynamic anchor assignment strategies. Typical approaches adaptively decide \posneg labels based on the proposal qualities on each anchor location using a customized detector likelihood \cite{zhang2019freeanchor}, the statistics of anchor IoUs \cite{2020atss}, or anchor scores \cite{ke2020mal,kim2020paa}. OTA \cite{ge2021ota} further considers that a better assignment strategy should be from a globally optimal perspective, 
rather than defining \posneg samples for each GT independently.
These strategies suggest that anchor positions with high-quality predictions should be assigned to related GTs, while anchors with uncertain predictions are labeled as negatives.  Other methods demonstrate that it is better top focus learning on the high-quality examples of objects
and thus they re-weight samples during training to control the contributions of different anchors by designed scores such as IoU-based weights \cite{wu2019ioubalance}, cleanliness scores \cite{li2020noisyanchor}, differentiable confidence \cite{zhu2020autoassign}, and IoU/score-ranking importance \cite{cao2020pisa}. However, these advanced strategies are not specially designed for crowded scenarios, and few are evaluated on the crowd datasets. \abc{Furthermore,  these two strategies, assignment and reweighting, are considered separately.}
 
In this paper, we consider a jointly optimal solution of anchor assignment and re-weighting in crowded scenes, where overlapping objects increases the number of ambiguous anchors.  Specifically, we propose a {\em unified framework for both anchor assignment and anchor re-weighting}.
To effectively assign anchors,  the confidences of anchor locations for each object need to be evaluated, reflecting the detector's ability to  consistently identify the object from that location.
To this end, we estimate the confidences (i.e., weights) 
of anchor locations summed over all objects as an {\em object density map} (Fig.\ref{fig:figure1}(b)), which is predicted from the image using an additional prediction layer.	
The density map predictor is supervised using an unbalanced optimal transport loss (UOTloss) between the predicted density map and the GT density mass (each object containing one unit density). The UOT defines a global optimization problem (over all GT objects and anchor locations), which assigns each GT's density (or part thereof) to anchor locations in the density map, where each (partial) assignment incurs a transport cost. 
\zcy{From the assigned individual density distribution of each object, we decode the anchor assignment and weighting results (Fig.\ref{fig:figure1}(c)).}
In UOT, we design an {\em overlap-aware transport cost}, which reduces ambiguous anchor assignments caused by neighboring objects with overlapping bounding boxes. Meanwhile, the transport cost is based on the prediction quality of each anchor location using the current training state, with higher quality predictions having lower transport cost.
In this way, the detector learns to use the optimal locations from which to classify and localize the objects.  
We denote our proposed anchor assignment and re-weighting strategy as Density-Guided Anchors (DGA). 

Moreover, the predicted density map is trained by UOTloss to present the summary weights over all objects for each anchor position in the image, which can naturally reflect the object density on each location. Thus, we take advantage of the density map, and design density-guided NMS (DG-NMS), which utilize predicted densities to adaptively adjust the NMS threshold to reduce the mistaken suppression in the post-processing.

%
The contributions of our paper are summarized as follows:
1) we propose a unified anchor assignment and re-weighting approach (DGA) especially for high-overlapping crowded object detection, which is based on a globally optimized transport plan matrix estimated from our UOTloss;
2) We propose an overlap-aware transport cost, which reduces ambiguous anchor assignments caused by neighboring objects with overlapping bounding boxes;
3) We design a novel density-guided NMS (DG-NMS), which helps to mitigate erroneous suppressions in crowded scenarios.
4) Comprehensive experiments on CrowdHuman \cite{shao2018crowdhuman} and  Citypersons \cite{zhang2017citypersons} demonstrate the effectiveness and generalizability of our DGA and DG-NMS.


\section{Related Work} \label{sec:related}

\subsection{Object detection in crowded scenes}
The difficulties of crowded object detection are mainly introduced by the high-overlapping and occlusion of objects. Some previous works \cite{lu2020semantic, chi2020relational, gahlert2020visibility, huang2020nms, zhang2019double} mitigate this problem by training with extra information, such as the bounding boxes of visible regions or human heads, which may include clearer cues of the objects. Then, the paired proposals of full (body) and visible (head) parts can jointly decide the final predictions. Other methods propose new loss functions for predicting stricter and tighter boxes, e.g., Aggregation \cite{zhang2018occlusion} and Repulsion Losses \cite{wang2018repulsion}, which encourage the proposals to be locating compactly around the ground truth. 
\zcy{Without using extra information, our DGA boosts the detection in crowded scenes via handling the ambiguous anchor problem by optimizing anchor assignment and re-weighting during training, which is rarely considered in previous works.}

Other approaches design more effective duplicate removal methods, since the key assumption of classical NMS does not hold when objects are partially occluded by neighboring objects. Soft-NMS \cite{bodla2017soft} uses the score decay to replace the hard deletion of neighboring proposals. \cite{hosang2017learning,qi2018sequential,hosang2016convnet} explore using neural network to perform the function of duplicate removal. Other methods employ training in the NMS process, such as predicting feature embeddings \cite{salscheider2021featurenms} to refine NMS, or multiple predictions with an anchor to perform set suppression \cite{chu2020detection}.
Adaptive-NMS \cite{liu2019adaptive} defines the object density as the max bounding box intersection over union (IoU) with other objects, and varies the NMS threshold based on predicted densities. Our DG-NMS also adjusts the NMS threshold based on object densities, \abc{but in contrast to  \cite{liu2019adaptive}}, our density map is trained by UOTloss according to the optimal density assignment result, which has no hand-crafted ground truth definition.

\subsection{Anchor assignment and re-weighting} 
Crowded object detection methods usually use the same anchor assignment strategies as their  object detection baselines during training. 
For anchor-based detectors, the classic approach thresholds the IoUs between anchors and GT to assign their \posneg labels, and has been widely adopted by Faster R-CNN based methods \cite{2015fasterrcnn,cai2018cascade,2017maskrcnn,lin2017fpn,zheng2020cross}, as well as some anchor-based one-stage detectors \cite{2017yolo9000,2018yolov3,2020yolov4,2016ssd,2017focal}.
For anchor-free detectors, which do not rely on anchor boxes, the spatial positions around the center of object boxes are generally regarded as positive anchor points \cite{2016yolo,kong2020foveabox,huang2015densebox,yu2016unitbox,2019fcos,2019center2}.

Recently, it has been found that the anchor assignment strategy is the key that causes the performance gap between anchor-based and anchor-free detectors \cite{2020atss}.
More flexible and adaptive anchor assignment strategies (e.g., MetaAnchor \cite{yang2018metaanchor}, GuidedAnchor \cite{wang2019guidedanchor}, ATSS \cite{2020atss}, PAA \cite{kim2020paa}) are proposed to eliminate the drawbacks of the fixed assignment based on IoU or central region. Other studies (e.g., NoisyAnchor \cite{li2020noisyanchor}, AutoAssign \cite{zhu2020autoassign}, PISA \cite{cao2020pisa} and IoU-balance \cite{wu2019ioubalance}) consider anchor re-weighting, designing weights for emphasizing the \zcy{training} 
samples on anchors that play a key role in driving the detection performance.

OTA \cite{ge2021ota} firstly models the anchor assignment as an optimal transport (OT) problem that transports positive labels from GTs to anchors. However, OTA assigns binary labels to anchors, and thus does not consider the importance weights of positives labels. Furthermore, OTA needs to estimate and fix the number of positive labels before solving the OT assignment. 
In contrast to OTA, we propose DGA which uses an unbalanced OT (UOT) problem to generate a density distribution for each object, from which we adaptively determine the number of positive anchors and their confidences (\abc{weights}). 
Thus, in contrast to OTA, we jointly and globally solve for the anchor assignment and re-weighting using UOT and predicted object density map. Moreover, aiming at address the problem of ambiguous anchors in crowded scenarios, we specifically design an overlap-aware transport cost for UOT.

\begin{figure*}[t]
	\begin{center}
		\includegraphics[width=0.85\linewidth]{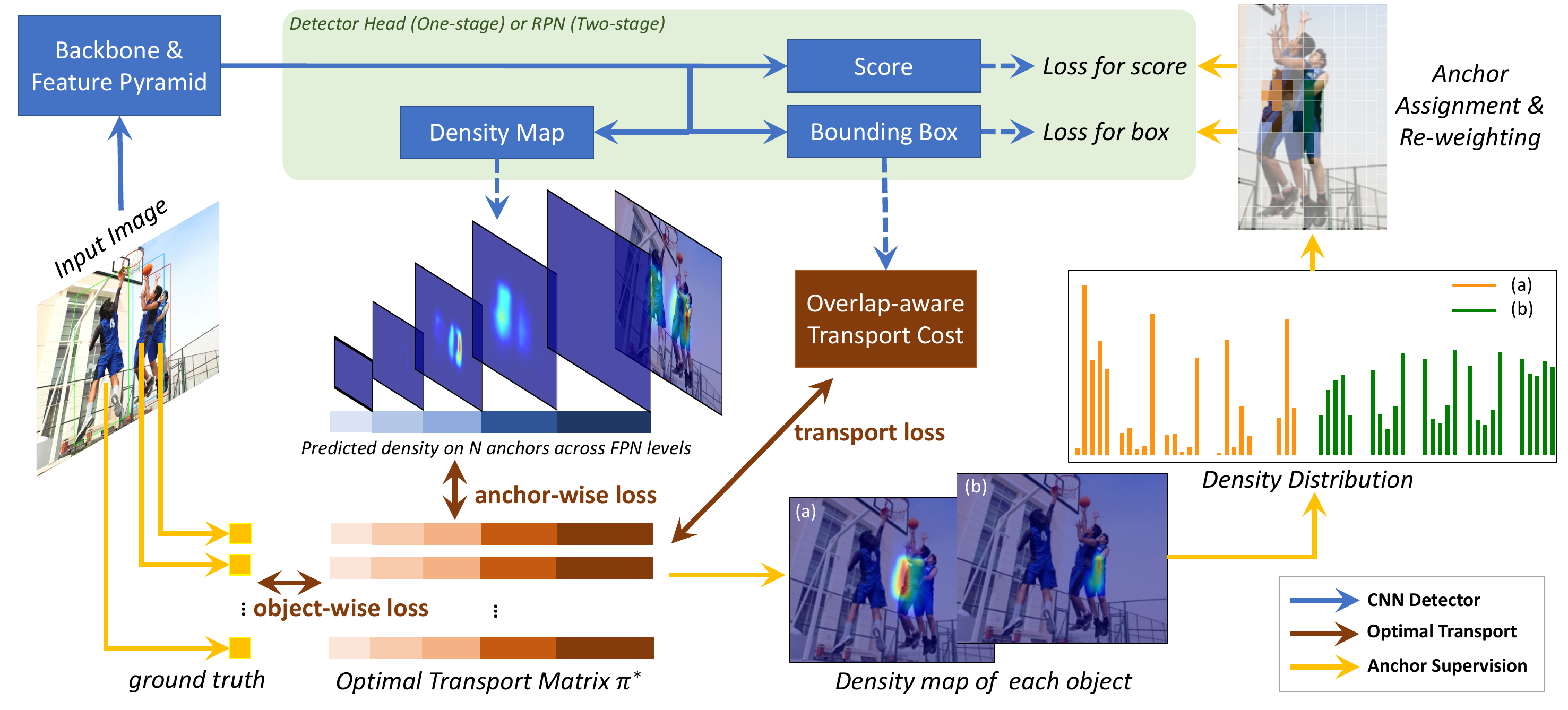}
	\end{center}

	\caption{The framework of our Density-Guided Anchors (DGA). Anchor confidence is represented by the predicted multi-level density map, which is learned through the proposed UOTLoss using unbalanced optimal transport (UOT).  The transport cost in UOT is based on the prediction quality of the anchor location, with better prediction quality yielding lower transport cost. 
		In this way, the detector learns the optimal locations from which to classify and localize the objects. The instance-wise density map for each object is extracted from the optimal transport plan matrix, and is then used to generate the anchor assignments and anchor weights.
	} 
	\label{fig:framework}

\end{figure*}

\section{Density-Guided Anchors}
\label{sec:dg-anchors}

In this section, we introduce the proposed Density-Guided Anchors (DGA), aiming to evaluate the confidence of anchors towards each GT from a global perspective, while jointly solving for  anchors assignment and re-weighting. DGA consists of three components: 1) a multi-level density map representing anchor confidence, which is globally optimized by UOTLoss; 2) density-guided anchor assignment; and 3) density-guided anchor re-weighting. The pipeline is presented in Fig.~\ref{fig:framework}.


\subsection{Density Map Prediction and UOTLoss}

The anchor confidence is represented as a multi-level density map, which sums to the number of objects.
Let there be $m$ GT objects and $n$ anchors across all FPN \cite{lin2017fpn} levels for an input image $I$.  
The density mass owned by each GT is set to one, denoted by $\mathbf{a}=[a_{i}]_{i=1}^{m}=\mathbf{1}_{m}$,
The density value predicted by the network for the $j$-th anchor location is $b_{j}$, and $\mathbf{b}=[b_{j}]_{j=1}^{n}$ is the flattened multi-level density map. 
To measure the loss between the GT and the predicted density map during training, we use an optimal transport problem, which aims to find the minimal cost for transporting the densities from the GTs to the anchors.
The transport cost matrix between GTs and anchors is $\mathbf{C} \in \real_{+}^{m \times n}$, whose entry $C_{ij}$ measures the cost of moving the density owned by the $i$-th object to the $j$-th anchor position. 
By defining $\mathbf{C}$ based on the prediction qualities of the anchors, with higher quality having lower cost (see next subsection), the model is supervised to predict higher density (higher confidence) over the high-quality anchors.

To be robust 
to outliers, we use the {\em unbalanced optimal transport} (UOT) problem \cite{peyre2019computational, feydy2019interpolating, wan2021generalized}, which allows some density from the GT or anchors to be unassigned,
\begin{equation}
		\footnotesize
		\mathcal{L}_{C}^{\varepsilon}(\mathbf{a}, \mathbf{b})=
		\min_{\bpi \geq 0} 
		\left \langle \mathbf{C},\bpi \right \rangle + 
		\varepsilon \mathrm{H} ( \bpi )  		+ \mathrm{KL}\left ( \bpi \mathbf{1},\mathbf{a} \right )+ \mathrm{KL}( \bpi^{\mathrm{\mathsf{T} } }\mathbf{1}, \mathbf{b}), 
	\label{equ:uot}
\end{equation}\normalsize
where $\bpi\in \real_+^{m \times n}$ is the transport plan matrix, where $\pi_{ij}$ indicates the density moved from the $i$-th GT to $j$-th anchor. In (\ref{equ:uot}), the first term is the total transport cost for the plan $\bpi$, while the third and fourth terms are the Kullback-Leibler (KL)
penalties for assigning only partial ground-truth density or partial anchor density, respectively. 
The 2nd term 
is the entropic regularization term ($H$ is the entropy function), which enables efficient solving on GPU\footnote{OT toolbox: https://www.kernel-operations.io/geomloss/index.html}. $\varepsilon$ is the hyperparameter controlling the effect of  the regularization.

The optimal transport plan matrix $\bpi^{*}$ is the plan associated with the minimal loss in (\ref{equ:uot}).
Summing over its rows yields the total transported object density, $\hat{\mathbf{a}}=\bpi^{*}\mathbf{1}$, i.e., the \abc{(fractional) number} of anchors that have been assigned to each GT object, while summing over its columns yields the reconstructed object density map, $\hat{\mathbf{b}}=\bpi^{*\mathsf{T}}\mathbf{1}$, i.e., the \abc{(fractional)} number of objects matched to each anchor. 
To further encourage all GT objects to be assigned to anchors and all anchors to be used in the predicted density map, we include two additional terms, giving our complete UOTLoss:

\begin{equation}
	\mathcal{L}^{uot}  = \mathcal{L}_{C}^{\varepsilon} + D_{1}(\mathbf{\hat{a}} , \mathbf{a} )+ D_{2}(\mathbf{\hat{b}} , \mathbf{b} ).
	\label{equ:complete_uot}
\end{equation}
The 1st term $\mathcal{L}_{C}^{\varepsilon}$ is the transport loss using the optimal transport plan $\bpi^*$ and cost matrix $\mathbf{C}$.
The 2nd term $D_{1}(\mathbf{\hat{a}} , \mathbf{a} ) $ is an object-wise loss to ensure that all density from each object is transported, i.e., all objects are assigned to predicted anchors.
The 3rd  term $D_{2}(\mathbf{\hat{b}} , \mathbf{b} )$ is an anchor-wise loss to ensure that all anchors with non-zero density are used, and that not too many anchors are predicted. In our work, $D_1$ is the Focal Loss \cite{2017focal} and $D_2$ is L2 norm.

\subsection{Overlap-aware Transport Cost}

The transport cost directly influences the density transported (assigned) from GTs to anchors, which should reflect the prediction quality of the anchor based on the current training state. 
Aiming at reducing the ambiguous anchors in crowded scenes, we propose an overlapped-aware cost function calculated with regression predictions in the candidate set and ground truths.



\subsubsection{Overlap-aware cost}
The overlap-aware cost measures the quality of the predicted bounding box (bbox) based on its overlap with a given GT bbox, and its distinctiveness (non-overlapping) from other GT bboxes.
Intuitively, for a good anchor, the IoU between its predicted bbox and corresponding GT should be near to 1.
Furthermore, to discourage ambiguous anchors that cover more than one object, the IoU of its predicted bbox to {\em other} GT should be near 0. 

\begin{figure}[t]
	\begin{center}
		\includegraphics[width=0.45\textwidth]{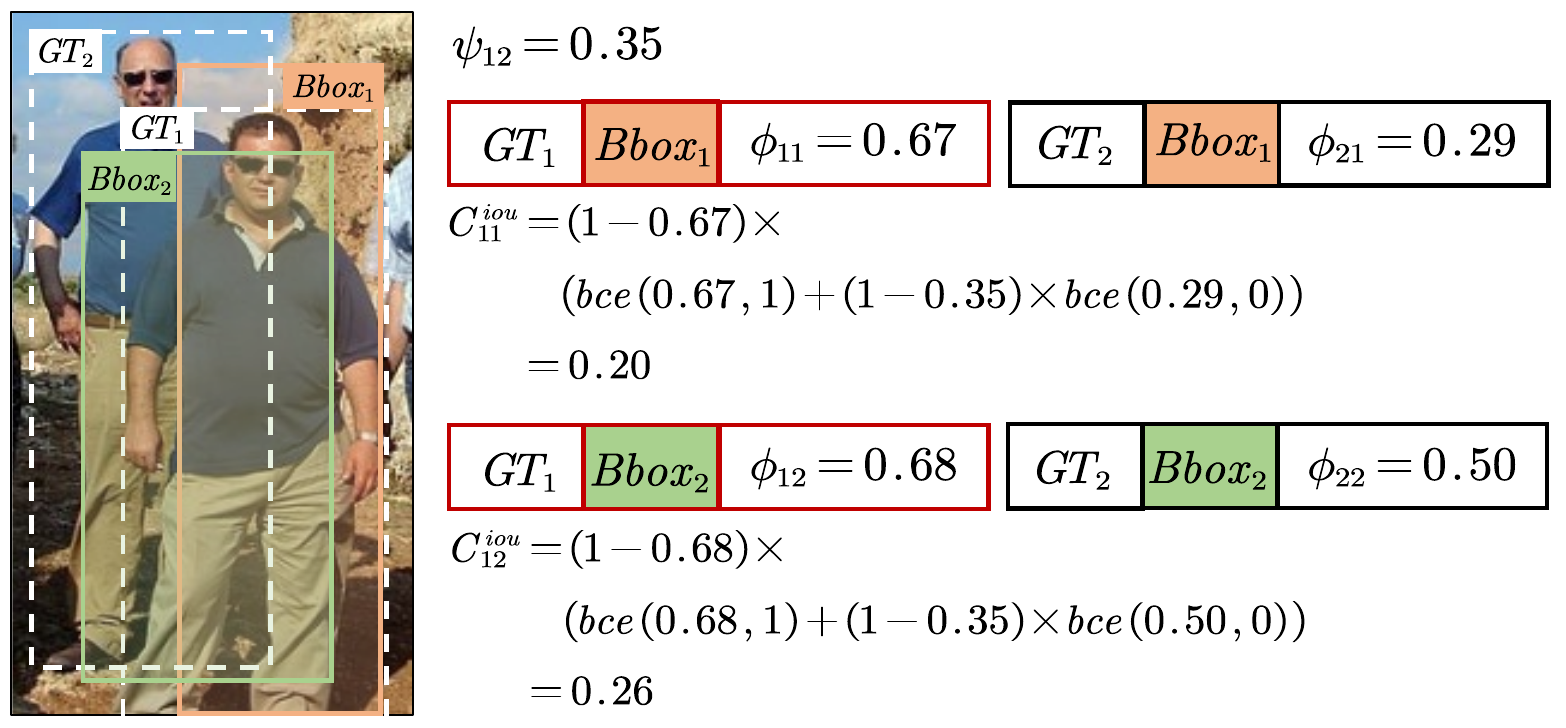}
	\end{center}

	\caption{An example of our overlap-aware cost.
		The predicted $Bbox_{1}$ and $Bbox_{2}$ have similar IoU with $GT_{1}$.
		However, $Bbox_{1}$ has less overlap with $GT_2$, and thus has less overlap-aware cost $C_{11}^{iou}$ compared to $C_{12}^{iou}$ of $Bbox_2$. Thus, the unbalanced optimal transport problem will prefer assigning $GT_1$ to $Bbox_1$.
	}
	\label{fig:costiou}
\end{figure}

Define the IoU between the $i$-th GT bbox and the $j$-th predicted bbox as $\phi_{ij}$, and define the IoU between the $i$-th GT bbox and the $k$-th GT bbox as ${\psi}_{ik}$, $k\neq i$. 
To measure the cost of assigning the $i$-th GT to the $j$-th predicted bbox, we could use the binary cross-entropy (BCE) loss on the IoUs between each GT and the $j$-th predicted bbox, and their ideal values (either 1 for the $i$th GT, or 0 otherwise), $\sum_{k=1}^m L_{bce}(\phi_{kj}, \indicator(k=i))$.
However, considering that another GT may be overlapped with the $i$-th GT, the corresponding loss $ L_{bce}\left(\phi_{kj}, 0 \right) $ is reduced based on its proximity to the $i$-th GT box $\psi_{ik}$. Then, the sum over all GTs is weighted by the IoU distance of the $(i, j)$ pair, so that the cost is zero when the prediction perfectly fits the GT:

\begin{equation} 
	\footnotesize
	C_{ij}^{iou}=\big(1-\phi_{ij}\big) \sum_{k=1}^{m} \left(1-{\psi}_{ik} \right)^{\indicator(k\neq i)} L_{bce}\left (\phi_{kj}, \indicator(k=i) \right ), 
	\label{equ:ioucost}
\end{equation}\normalsize
where $\indicator(\cdot)$ is the indicator function.
This design works well in situations where multiple objects are overlapping since \abc{each predicted bbox is encouraged to focus on one GT, while being distinctive from other GTs} (e.g., see Fig.~\ref{fig:costiou}).

\subsubsection{Candidate priors} The previous anchor assignment methods \cite{zhang2019freeanchor,2020atss,kim2020paa,ge2021ota} uses anchors around the object centers as candidates to be assigned. We also adopt the similar strategies based on baselines.
For \emph{anchor-free} detector frameworks like FCOS \cite{2019fcos}, the {\em Center Prior} setting is also used, which selects the candidate anchor points for each object as the $r^2$ closest anchors from each FPN level according to the distance between anchor points and the center of objects. 
\zcy{The selected anchors are involved in the UOT optimization and assignment process, while others are regarded as negative (zero-density) locations \abc{and ignored}.}
Based on the design purpose of FPN where objects of different sizes are better predicted at different scale levels, we further define a {\em Level Cost} instead of the hard assigning GTs to levels. For each GT, one or two preferred FPN levels, denoted as $L_i$, are assigned based on the GT's size and each level's SoI (Size of Interest). The smallest level difference between the $i$th GT and $j$th anchor is defined as the level cost,

\begin{align}
	C_{ij}^{level} = \min_{a \in L_i}  |a-l_j|,
\end{align}
where $L_i$ is the set of preferred FPN levels for the $i$-th GT, and $l_j$ is the level of the $j$th anchor.  The final transport cost is $C_{ij} = \gamma C_{ij}^{iou} + C_{ij}^{level}$, where we set $\gamma=2$.

For \emph{anchor-based} frameworks, anchors whose IoU with any GT larger than a threshold are regarded as candidates, which is inherited from the baselines such as one-stage RetinaNet \cite{2017focal} and the RPN in two-stage FPN \cite{lin2017fpn}.
The candidate set increases the attention on potential positive areas during UOT, which give a reliable indicator in the early training and help to stabilize the training process. 

The influence of {\em Center Prior} $r$ and {\em Level Cost} are evaluated in Sec.~\ref{sec:ablation}, and we adopt $r=5$ with level cost 
in the experiments for anchor-free detectors. For anchor-based methods, the IoU threshold for 
\zcy{obtaining the} anchor candidates is consistent with the corresponding baselines. 
We also report the effect of this threshold in the supplemental.

\begin{figure}[t]
	\begin{center}
		\includegraphics[width=0.45\textwidth]{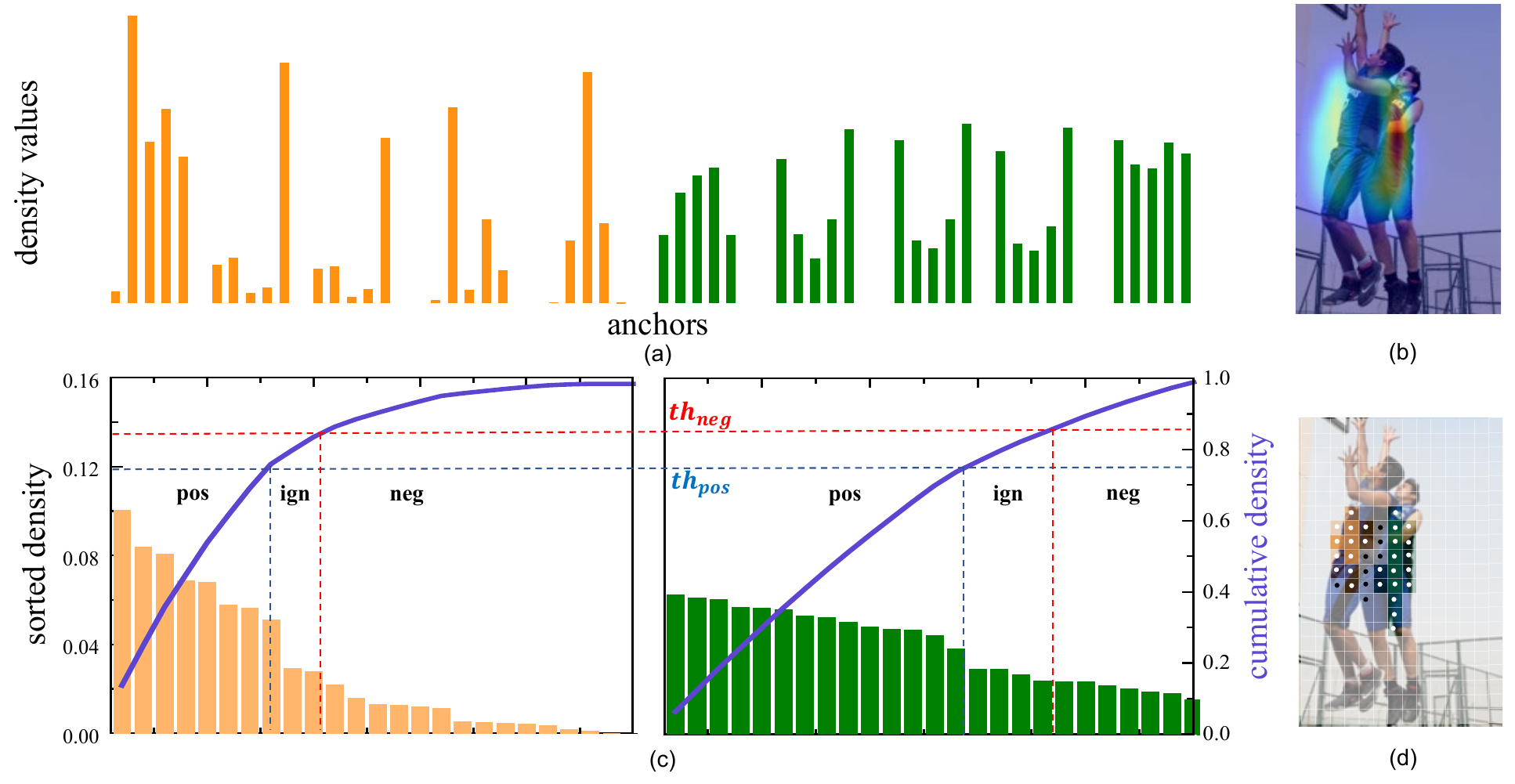}
	\end{center}
	\caption{The anchor assignment process in DGA. 
		(a) Transported density values of the two objects over the anchor positions, corresponding to the values in two rows of the OT plan $\bpi^*$.
		(b) The transported density on anchors reshaped to a map, and visualized on the image. 
		(c) The sorted density values for an object (bar plot), the cumulative  density (bold line), and the partition of \posneg and ignore (\emph{ign}) labels based on the positive and negative thresholds, $th_{pos}$ and $th_{neg}$. (d) The positive labels and ignored labels visualized on the image as white and black circles, respectively.
	}
	\label{fig:assignment}
\end{figure}

\subsection{Anchor Assignment}
The OT plan $\bpi^{*} \in \mathbb{R} ^{m \times n}$ contains the density assignment from $m$ GT to $n$ anchors. The matrix can be observed from two views:  1) each row of $\bpi^*$ is the density distribution of each GT, which indicates the anchor confidence with respect to a specific object (see Fig.~\ref{fig:assignment}a and \ref{fig:assignment}b); 2) each column of $\bpi^*$ represents the density mass obtained from each GT on the same anchor, which indicates the
competition of the GTs for one anchor. 

We dynamically assign positive labels to anchors with high density values (the ``head'' of the density distribution), and negative labels to the anchors with low density values (the ``tails''). Specifically, for each object, we sort its corresponding instance-level density values (its row in $\bpi^*$) in descending order and compute a cumulative sum.
We then assign  \posneg labels to the anchors based on thresholds on the cumulative density mass (see Fig.~\ref{fig:assignment}c).  The anchors corresponding to the accumulated mass below the threshold $th_{pos}$ are assigned positive labels, while the anchors for accumulated mass above a threshold $th_{neg}$ are assigned negative labels. The remaining anchors are ignored (see Fig.~\ref{fig:assignment}d).
For the case where multiple GTs compete for the same anchor, the GT with largest value in the anchor's column of $\bpi^*$ is selected as positive, while others are negative.
In this way, the confident anchors will be involved, and the number of positives for each GT is adaptively decided by the instance-level density distribution, which is globally optimized and related to the current training state.

\subsection{Anchor Re-weighting and Final Loss}

For anchor re-weighting, the importance of the positive samples for a GT can be ranked based on their density values.
The anchor weights for a positive GT sample are obtained by normalizing its corresponding density values so that the most confident anchor has weight of 1. Specifically, $w_j = d_j / \max_j d_j$, where $\{d_j\}_j$ are the density values of the positive samples of the $i$-th GT.
Note that all negative samples are equally weighted without special design. 
With the proposed sample re-weighting, the classification loss and localization loss  are

\begin{align}
	\mathcal{L}^{cls} =& \tfrac{1}{\sum_j w_{j}} \big ( \sum\nolimits_{j\in {\cal A}^{p}}w_{j}\mathcal{L}_{j}^{cls} +\sum\nolimits_{j\in {\cal A}^{n}}\mathcal{L}_{j}^{cls} \big), \\
	\mathcal{L}^{loc} =& \tfrac{1}{\sum_j w_{j}}\sum\nolimits_{j\in {\cal A}^{p}}w_{j}\mathcal{L}_{j}^{reg}, 
\end{align}
where ${\cal A}^{p}$ and ${\cal A}^{n}$ represent the sets of positive and negative anchors.
We adopt GIoU Loss for $\mathcal{L}^{reg}$ and Focal Loss for $\mathcal{L}^{cls}$. With focal loss weighted by DGA, the classification training comprises both hard samples and important samples for positives. 

With DGA, both the \posneg anchor assignment  and re-weighting are decoded from the anchor confidence represented in the optimal transport plan $\bpi^{*}$. 
Note that the CNN that predicts the density map is also updated with $\bpi^*$ through $\mathcal{L}^{OT}$. 
Finally, the three key processes in detector training: anchor confidence estimation, anchor assignment, and re-weighting, are solved consistently in  end-to-end training, where the final loss function is

\begin{equation}
	\mathcal{L}=\mathcal{L}^{cls} + \gamma_{1} \mathcal{L}^{loc} + \gamma_{2} \mathcal{L}^{uot}, 
	\label{equ:finalloss}
\end{equation}
where $\gamma_1$ is consistent with baseline framework ($\gamma_1=\{2,1\}$ for \{one-stage, two-stage\} detectors), $\gamma_2$ is empirically set to 0.25 based on the range of UOTLoss values.

\section{Density-Guided NMS (DG-NMS)}
\label{sec:dg-nms}

The training of density map prediction is supervised by the UOTLoss in (\ref{equ:complete_uot}), where the GT is the summation over columns of the optimal transport plan. Thus, the density map learns from the summary of densities assigned from each object, which is able to reflect the local degree of crowdedness. 

The vanilla NMS uses a 
\zcy{fixed}
threshold to suppress duplicates; a higher threshold yields more false positives at low-density areas, while a lower one results in missing highly-overlapped objects. We propose a density-guided NMS, which adaptively adjusts the threshold $T_{i}$ for the $i$-th proposal box $b_{i}$ based on the predicted density $d_{i}$, via:

\begin{equation}
	T_{i} = 0.5 + 0.3f(d_{i}),
	\label{equ:nms_thresh}
\end{equation}
\zcy{where $f(\cdot)$ is a min-max scaling function that maps the density values for an image to be between 0 and 1.
Therefore, the threshold $T_i$ is adjusted linearly between 0.5 and 0.8 according to the local  density.}

\zcy{In the NMS procedure, starting with a set of proposals $\mathcal{P}$, the prediction $b_{t}$ with the maximum score is selected and put into the set of final predictions $\mathcal{F}$. Then with the $b_{t}$ as a target, any box in $\mathcal{P}$ that has overlap with $b_{t}$ higher than a threshold $T$ is removed. 
In our DG-NMS, we use the adaptive $T_{i}$ to consider keeping/removing each  $b_{i}$ in $\mathcal{P}$, where
box $b_{i}$ is \zcy{kept} when $IoU(b_{t}, b_{i}) \leq T_{i}$.}
\abcn{As boxes are removed, the remaining boxes kept in $\mathcal{P}$ become less crowded, and thus we correspondingly reduce the density values for the kept boxes $b_i$,}

\begin{equation}
	d_{i} \leftarrow d_{i}\cdot\exp(-{IoU(b_{i},b_{t})^{2}}/{\sigma}),
	\label{equ:d_decay}
\end{equation}
which is based on the IoU of the kept $b_i$ with the current target box $b_t$ \abcn{(higher IoU yields larger decay).}
\zcy{The duplicates removal process and density reduction step is repeated for the new maximum-score box as $b_t$, until $\mathcal{P}$ becomes empty.}
The pseudo code of our DG-NMS is in the supplemental.
The ablation studies about $f(\cdot)$ and $\sigma$ are shown in Tab.~\ref{tab:dg-nms}.

Through this design and the help of the predicted density, the density-guided NMS keep more predictions ub highly-overlapped regions with higher NMS thresholds, and fewer false positives at low-density regions with lower thresholds.

\section{Experiments} \label{exp}

In this section, we conduct extensive experiments to evaluate the proposed approach from different aspects. To verify its effectiveness on crowd scenes, we adopt the typical crowded datasets, CrowdHuman \cite{shao2018crowdhuman} containing an average of 22.64 (2.40) objects (overlaps) per image, and Citypersons \cite{zhang2017citypersons} containing an average of 6.47 (0.32) objects (overlaps) per image. Most of the comparisons and ablation experiments are performed on the heavily crowded CrowdHuman dataset, while the results on moderately crowded Citypersons dataset are reported to demonstrate the generality and robustness in crowded scenarios. 


\textbf{Evaluation metrics}
Following \cite{chu2020detection}, we mainly adopt three evaluation criteria: Average Precision (AP$_{50}$); Log-Average Missing Rate (MR), commonly used in pedestrian detection, which is sensitive to false positives (FPs) in high-confidence predictions; Jaccard Index (JI). Generally, larger AP$_{50}$, JI and smaller MR indicate better performance.

\textbf{Implementation Details}
The proposed DGA can be adopted by both anchor-based and anchor-free detectors. The density prediction and the process of anchor assignment are performed at the detector head for one-stage detector, and at the RPN for two-stage detector. The density map regressor is implemented as one convolutional layer on top of the bbox regression head.
We compared with baselines for multiple detectors, such as one-stage anchor-free FCOS \cite{2019fcos}, one-stage anchor-based RetinaNet \cite{2017focal}, and two-stage FPN \cite{ghiasi2019nasfpn}.
The ablation studies are mainly reported with the framework of FCOS \cite{2019fcos} due to its simplicity. For related implementations of FCOS, an auxiliary IoU branch is adopted as a default component similar to recent one-stage detectors \cite{2016yolo,2019fcos,kim2020paa,ge2021ota}.
Unless otherwise specified, the default hyper-parameters used in the baselines are adopted, and the training protocol is consistent as in \cite{chu2020detection}. The batch size is 16, with mini-batch is 2. Multi-scale training and test are not applied, and the short edge of each image is resized to 800 pixels for both train and test. 
We utilize standard ResNet-50 \cite{he2016resnet50} pre-trained on ImageNet \cite{deng2009imagenet} as the backbone network. With density prediction in our approach, one anchor point/box for one location is used. For the anchor box setting, the aspect ratio is set to H:W=3:1 for both CrowdHuman and Citypersons. 

In UOT of (\ref{equ:uot}), we set $\varepsilon=0.7$ for anchor-free detectors and $\varepsilon=0.1$ for anchor-based ones (see supplemental about the ablation study of $\varepsilon$).
In DGA, $th_{pos}=0.7$ and $th_{neg}=0.8$. \zcy{For DG-NMS, $\sigma=0.5$,}
which is shown to be insensitive on results in Tab.~\ref{tab:dg-nms}.

\subsection{Comparisons with different methods}

\begin{table}
	\setlength\tabcolsep{0.2in}
	\centering
	\caption{Comparisons on CrowdHuman validation set. $\dagger$ indicates the method is implemented by PBM \cite{huang2020nms}.}
	\begin{tabular}{lccc}
		\hline
		Method  & AP$_{50}\uparrow$ & MR$\downarrow$ & JI$\uparrow$ \\
		\hline		
		\hline
		{\em One-stage anchor-free} \\
		FCOS \cite{2019fcos} & 86.8 & 54.0 & 75.7 \\
		FreeAnchor \cite{zhang2019freeanchor} & 83.9 & 51.3 & - \\
		PAA \cite{kim2020paa} & 86.0 & 52.2 & - \\
		POTO \cite{wang2021end} & 89.1 & 47.8 & 79.3 \\
		OTA \cite{ge2021ota} & 88.4 & 46.6 & - \\
		FCOS+DGA & 88.3 & 44.5 & 78.6 \\
		\hspace{0.5cm}+DG-NMS & \textbf{89.5} & \textbf{44.1} & \textbf{80.7} \\
		\hline
		\hline
		{\em One-stage anchor-based} \\
		RetinaNet \cite{2017focal} & 85.3 & 55.1 & 73.7 \\
		ATSS \cite{2020atss} & 87.0 & 51.1 & 75.9 \\
		RetinaNet+DGA & 87.5 & \textbf{48.4} & 77.3 \\
		\hspace{0.5cm}+DG-NMS & \textbf{88.4} & 48.6 & \textbf{78.8}\\
		\hline
		\hline
		{\em Two-stage} \\
		FPN$^{\dagger}$ & 84.9 & 46.3 & - \\
		Adaptive NMS$^{\dagger}$ \cite{liu2019adaptive} & 84.7 & 47.7 & - \\
		PBM$^{\dagger}$ \cite{huang2020nms} & 89.3 & 43.4 & - \\
		\hline
		FPN \cite{lin2017fpn} & 85.8 & 42.9 & 79.8 \\
		RelationNet \cite{hu2018relation} & 81.6 & 48.2 & 74.6 \\
		GossipNet \cite{hosang2017learning} & 80.4 & 49.4 & 81.6 \\
		FPN+MIP \cite{chu2020detection} & 90.7 & 41.4 & 82.4 \\
		FPN+DGA & 89.5 & 41.7 & 80.4 \\
		\hspace{0.5cm}+DG-NMS & 90.4 & 41.9 & 81.7 \\
		FPN+MIP+DGA & 91.9 & 41.3 & 82.2 \\
		\hspace{0.5cm}+DG-NMS & \textbf{92.2} & \textbf{41.1} & \textbf{82.6} \\
		\hline		
	\end{tabular}

	\label{tab:main}
\end{table}
In CrowdHuman, there are 15k, 4.37k, and 5k images in the training, validation, and test sets. As is common practice \cite{chi2020relational,liu2019adaptive, zhang2019double, huang2020nms, chu2020detection}, we report results of human full-body on the validation set.
We compare with different types of mainstream detectors, including one-stage anchor-free \cite{2019fcos,wang2021end,ge2021ota}, anchor-based \cite{2017focal, 2020atss}, and two-stage \cite{lin2017fpn, liu2019adaptive, chu2020detection, huang2020nms} approaches. As shown in Tab.~\ref{tab:main}, by equipping with our DGA, significant performance improvements have been achieved on both one-stage and two-stage baselines, illustrating the effectiveness of our anchor assignment and re-weighting approach in handling crowded scenes. Concretely, DGA exhibits substantial gains over FCOS with improvements of 1.5 AP, 9.5 MR, and 2.9 JI; RetinaNet with improvements of 2.2 AP, 6.7 MR, and 3.6 JI; FPN with improvements 3.7 AP, 1.2 MR, and 0.6 JI. Furthermore, our DG-NMS consistently improves AP and JI around 1 to 2 with a comparable MR. Specifically, our DGA and DG-NMS is compatible with the current  state-of-the-art MIP \cite{chu2020detection}, which uses set-NMS. Through upgrading the anchor assignment strategy of MIP and combining DG-NMS with set-NMS, we obtain improvements of 1.5 AP, 0.3 MR and 0.2 JI.

\subsection{Ablation studies and analysis} \label{sec:ablation}

\begin{table}
	\setlength\tabcolsep{2pt}
	\centering
	\renewcommand\arraystretch{1.1}
	\caption{Ablation study of the UOTLoss on CrowdHuman valset.}
	\begin{tabular}{cccc|ccc}
		\hline
		\multicolumn{2}{c}{transport loss} & object-wise & anchor-wise & & &\\
		$\mathcal{L}_{C}^{\varepsilon}$ & $\left \langle \mathbf{C},\bpi^{*}  \right \rangle$ &  $D_{1}(\mathbf{\hat{a}}, \mathbf{a})$ & $D_{2}(\mathbf{\hat{b}},\mathbf{b} )$ & AP$_{50}\uparrow$ & MR$\downarrow$ & JI$\uparrow$    \\ \hline
		$\surd$ &      & 	     &		   & 87.5  & 45.7 & 78.8     \\
		$\surd$ &      & $\surd$ &	 	   & 87.7  & 45.5 & 78.3    \\
		$\surd$ &      &         & $\surd$ & 88.2  & 45.0 & \textbf{78.9}    \\ 
		$\surd$ &      & $\surd$ & $\surd$ & \textbf{88.3}  & \textbf{44.5} & 78.6      \\
		& $\surd$ & $\surd$ & $\surd$  & 87.8  & 45.7 &  78.7     \\
		\hline
	\end{tabular}
	\label{tab:uotloss}
\end{table}

\subsubsection{Effects of terms in UOTLoss.}  
We evaluate the effect of different terms in the proposed UOTLoss function in Tab.~\ref{tab:uotloss}. 
$\mathcal{L}_{C}^{\varepsilon}$ is the optimal solution of the UOT problem in (\ref{equ:uot}),  which includes entropic regularization and the marginal constraints, 
while $\left \langle \mathbf{C},\bpi^{*} \right \rangle$ is only the density transport cost using the optimal transport plan $\bpi^*$.
We compare these two transport losses and find that the former is superior, and applying extra penalties with object-wise loss $D_{1}$ and anchor-wise loss $D_{2}$ will both boost the performance.

\subsubsection{Effects of terms in overlap-aware transport cost.} We evaluate the effect of each component in the overlap-aware transport cost used in the UOT problem. Here we use Focal loss and IoU loss for $L_{score}$ and $L_{reg}$, respectively.   
The results are presented in the Table \ref{tab:transport_cost}. 
\zcy{\cite{kim2020paa,ge2021ota} use both predicted score and bbox to evaluate the quality of predictions. We first attempt to add $L_{score}$ to both $L_{reg}$ and our overlap-aware cost, but including the score prediction yields does not benefit  the performance. Thus, only box prediction results are involved in the overlap-aware cost, through which the UOT find the optimal assignment, and then supervises the score prediction branch with the consistent label.}

Compared with the standard IoU loss $L_{reg}$, using the overlap-aware cost significantly improves performance (AP 88.3 vs.~86.2; MR 44.5 vs.~46.7; JI 78.6 vs.~77.5).
In the case of anchor-free detector like FCOS, we also evaluate the effect of {\em Level Cost} and the radius $r$ for {\em Center Prior}. The result shows that using the {\em Level Cost} significantly improves the performance compared to without it (AP 88.3 vs.~87.1; MR 44.5 vs.~45.3).
Note that the level cost here is not a hard assignment that fixes the level of objects like FCOS,  but instead maps the difference between anchor's level and GT's pre-defined level as cost values in the transport cost. With this {\em soft} level prior and the overlap-aware transport cost defined in (\ref{equ:ioucost}), the UOT can obtain reasonable solutions when there are many close and overlapped targets.

The radius $r$ for {\em Center Prior} decides the number of candidate anchors for each GT. When $r \in \{3,5,7\}$, each object's corresponding numbers of candidate anchors are 45, 125, and 245 ($r^2$ times 5 FPN levels).
When $r \in \{3,5\}$, DGA achieves relatively better performance (AP of 88.7 and 88.3, MR 44.9 and 44.5, JI 79.1 and 78.6). When $r=7$ and more 
candidates are accounted for each GT, the AP performance drops 1.5 compared with $r=5$, which indicates that an appropriate positive candidate set is necessary.

\begin{table}
	\renewcommand\arraystretch{1.1}
	\centering
	\caption{Effect of terms in our overlap-aware transport cost (OA Cost) on CrowdHuman validation set. ``Center P.'' and  ``Level C.'' refer to center prior and level cost, respectively.}
	\begin{tabular}{c|c|c|ccc}
		\hline
		Transport Cost & Center P. & Level C.  & AP$_{50}\uparrow$ & MR$\downarrow$ & JI$\uparrow$ \\ \hline
		$L_{reg}+L_{score}$  & r=5  &  $\surd$  & 86.1 & 47.0 & 77.2 \\
		$L_{reg}$      & r=5  &  $\surd$  & 86.2 & 46.7 & 77.5 \\ \hline \hline
		OA Cost+$L_{score}$ & r=5  &  $\surd$  & 86.4 & 46.4 & 78.2\\ \hline
		               & r=3  &  $\surd$  & \textbf{88.7} & 44.9 & \textbf{79.1} \\
		OA Cost        & r=5  &           & 87.1 & 45.3 & 78.8 \\
		            & r=5  &  $\surd$  & 88.3 & \textbf{44.5} & 78.6 \\
		               & r=7  &  $\surd$  & 86.8 & 45.6 & 78.4 \\ \hline 
	\end{tabular}

	\label{tab:transport_cost}

\end{table}

\subsubsection{Comparison of anchor assignment strategies.}
Different from OTA \cite{ge2021ota}, which pre-defines the number of positive labels for each GT before solving the OT problem, we dynamically assign the number of positive labels based on the optimal transport plan.
Here we compare different anchor \posneg assignments:
``Dyn.~$k*$'' is our cumulative density method shown in Fig.~\ref{fig:assignment}, which uses two thresholds $th_{pos}$ and $th_{neg}$ to determine the \posneg labels.
``Dyn.~$k$'' is the strategy used in \cite{ge2021ota}, which uses the sum of IoUs between the top-20 anchors and GTs as the number of positives. 
``Fix.~$k$'' chooses the fixed number of top $k$ anchors for each object as positive candidates, while those whose density equals zero will be filtered out.
The Dyn.~$k$ strategy with the sum of IoU can also be interpreted as estimating the number of key anchors, while the assigned density in DGA gives a direct measurement.

The results are presented in Table \ref{tab:assignment}.
The $th_{pos}/th_{neg}$ settings of $70\%/70\%$ and $80\%/80\%$ obtain better AP and MR respectively, and we find that 
including a set of ``ignore'' anchors, where  $th_{pos}/th_{neg}$ is $70\%/80\%$, will further improve the AP by about 0.7 compared with  $70\%/70\%$ and MR by about 0.2 compared with $80\%/80\%$. 
Although Fix.~$k$ and Dyn.~$k$ seem to perform well on AP and JI, the MR is much worse (Fix.~$k$ 45.7, Dyn.$k$ 46.3 vs.~Dyn.~$k*$ 44.5), which indicates that the detector generates better high-confidence predictions with DGA, benefitting from the effective selection and assignment of key anchors. 

Furthermore, we also evaluate the influence of anchor re-weighing by removing it and setting all weights for positives anchors to one. The drops of AP and MR indicate the effectiveness of our weights to control the contributions of positive samples during training based on their qualities.

\begin{table}
	\renewcommand\arraystretch{1.1}
	\centering
	\caption{Comparison of different \posneg assignment strategies on the CrowdHuman validation set: (Dyn.~k*) thresholding the accumulated density mass (our strategy in Fig.~\ref{fig:assignment}); (Dyn.~k) using the sum of IoUs between the top-20 anchors and GTs as the number of positives (strategy in OTA \cite{ge2021ota}); (Fix.~k) fixed number of positives. AncR indicates using anchor re-weighting.}

		\begin{tabular}{ccc|c|ccc}
			\hline
			\multicolumn{3}{c|}{Anchor Assignment}  & AncR & AP$_{50}\uparrow$ & MR$\downarrow$ & JI$\uparrow$           \\ \hline
			\multicolumn{1}{c|}{FCOS} & \multicolumn{2}{c|}{Object Center} &   & 86.8 & 54.0 & 75.7\\ \hline
			\multicolumn{1}{c|}{}  & \multicolumn{1}{c|}{$th_{pos}$}  & $th_{neg}$ & &   &   & \\ 
			\multicolumn{1}{c|}{}  & \multicolumn{1}{c|}{0.7}  &  \multicolumn{1}{c|}{0.7} & $\surd$  & 87.6  & 45.2 & 78.5  \\
			\multicolumn{1}{c|}{Dyn. $k*$} & \multicolumn{1}{c|}{0.8}& \multicolumn{1}{c|}{0.8} & $\surd$ & 87.3 & 44.7 & 78.7 \\
			\multicolumn{1}{c|}{(Fig. \ref{fig:assignment})} & \multicolumn{1}{c|}{0.9}  & \multicolumn{1}{c|}{0.9} & $\surd$ & 86.7 & 45.1  & 78.6  \\
			\multicolumn{1}{c|}{} & \multicolumn{1}{c|}{0.7} &  \multicolumn{1}{c|}{0.8} & $\surd$ & \textbf{88.3} & \textbf{44.5} & 78.6 \\  
			\multicolumn{1}{c|}{} & \multicolumn{1}{c|}{0.7} &  \multicolumn{1}{c|}{0.8} & & 87.5 & 45.1 & \textbf{78.9} \\\hline
			\multicolumn{1}{l|}{Dyn. $k$}  & \multicolumn{2}{c|}{Sum of IoU}  & $\surd$ & \multicolumn{1}{c}{86.9} & \multicolumn{1}{c}{46.3} & \multicolumn{1}{c}{78.5} \\ \hline
			\multicolumn{1}{c|}{}    & \multicolumn{2}{c|}{Top5}  & $\surd$ & \multicolumn{1}{c}{87.6} & \multicolumn{1}{c}{45.7} & \multicolumn{1}{c}{78.1} \\
			\multicolumn{1}{c|}{Fix.~$k$}       & \multicolumn{2}{c|}{Top10}  & $\surd$ & \multicolumn{1}{c}{87.8} & \multicolumn{1}{c}{45.9} & \multicolumn{1}{c}{78.8} \\
			\multicolumn{1}{l|}{}       & \multicolumn{2}{c|}{Top15}  & $\surd$ & \multicolumn{1}{c}{87.4} & \multicolumn{1}{c}{46.2} & \multicolumn{1}{c}{78.5} \\
			\multicolumn{1}{l|}{}    	& \multicolumn{2}{c|}{Top20}  & $\surd$ & \multicolumn{1}{c}{86.9} & \multicolumn{1}{c}{47.1} & \multicolumn{1}{c}{78.3} \\ \hline
		\end{tabular}

	\label{tab:assignment}
\end{table}

\subsubsection{Effects of functions in DG-NMS.}
In Tab.~\ref{tab:dg-nms}, we evaluate  DG-NMS with different $f$ in (\ref{equ:nms_thresh}), including $f(d)=s(d)^2$, $f(d)=s(d)$, $f(d)=\sqrt{s(d)}$, where $s(\cdot)$ is the min-max scaling operation.
At the low density region, $s(d)^{2}$ generates low NMS thresholds, while $\sqrt{s(d)}$ generates relatively high NMS threshold, so that more proposals can be preserved by $\sqrt{s(d)}$.
We observe that there is a balance between AP and MR, since more detected instances will boost AP by increasing the total number of true positives, while also introducing more high-score false positives, which makes the MR worse. Therefore, we choose to adopt the simple $f(d)=s(d)$ in the experiment, which obtains the balanced results between using $s(d)^2$ and $\sqrt{s(d)}$.

The results with different $\sigma$ used for density decay in (\ref{equ:d_decay}) are also shown in Tab.~\ref{tab:dg-nms}. Note that the performance is not sensitive to the density decay speed. When choosing a $\sigma$ between $0.1$ to $1.0$, similar performance can be obtained.

\begin{figure}
	\begin{minipage}{0.26\textwidth}	
		\setlength\tabcolsep{2.2pt}
		\footnotesize
		\renewcommand\arraystretch{1.1}
		\centering
		\captionof{table}{Effectiveness of components in DG-NMS on {\em CrowdHuman val} set.}
		\begin{tabular}{cc|ccc}
			\hline
			$f(\cdot)$ & $\sigma$  & AP$_{50}\uparrow$ & MR$\downarrow$ & JI$\uparrow$ \\ \hline
			-         &  -   & 88.3 & 44.5 & 78.6 \\ \hline
			$s(d)^2$     &  0.5 & 89.0 & \textbf{44.0} & 80.1 \\
			$s(d)$       &  0.5 & 89.6 & 44.1 & 80.7 \\
			$\sqrt{s(d)}$&  0.5 & \textbf{90.2} & 44.7 & \textbf{80.8} \\ 
			 \hline \hline
			$s(d)$&  0.1 & 89.4 & 44.2 & 80.7 \\
			$s(d)$&  0.5 & \textbf{89.6} & \textbf{44.1} & \textbf{80.7} \\
			$s(d)$&  1.0 & 89.6 & 44.1 & 80.6 \\
			\hline		
		\end{tabular}
		
		\label{tab:dg-nms}
	\end{minipage}
	\hspace{0.1cm}
	\begin{minipage}{0.2\textwidth}
		\setlength\tabcolsep{2.2pt}
		\footnotesize
		\renewcommand\arraystretch{1.1}
		\centering
		\captionof{table}{Performance comparisons on {\em CityPersons val} set.}
		\begin{tabular}{l|cc}
			\hline
			Method & AP$_{50}\uparrow$ & MR$\downarrow$ \\ \hline
			RetinaNet \cite{2017focal} & 95.6 & 13.2 \\
			FCOS \cite{2019fcos} & 96.0 & 13.8 \\
			FPN \cite{lin2017fpn} & 95.2 & 11.7 \\
			FPN+MIP \cite{chu2020detection} & 96.1 & 10.7 \\ \hline
			RetinaNet+Ours & 96.3 & 12.9 \\ 
			FCOS+Ours & 96.3 & 12.0 \\
			FPN+Ours & \textbf{96.7} & \textbf{10.0} \\ 
			\hline		
		\end{tabular}

		\label{tab:citypersons}
	\end{minipage}
\end{figure}

\subsection{Experiments on CityPersons}

CityPersons \cite{zhang2017citypersons} contains 5000 images with size of $1024 \times 2048$, including 2975 for training, 500 for validation and 1525 for testing. Following the previous works \cite{zhang2017citypersons, chu2020detection}, the images in {\em reasonable} subset of training set and validation set are used with $1.3\times$ resolution compared to original ones for detector training and evaluation, respectively. We train all the models with 12 epochs and other hyperparameters are the same as before.

We adopt RetinaNet, FCOS, and FPN as baselines, and report our results in Tab.~\ref{tab:citypersons} using both DGA and DG-NMS. Our approach boosts the performance over all the different baselines on both AP and MR. Moreover, FPN with our approach surpasses the state-of-the-art MIP by increasing 0.6 AP and decreasing 0.7 MR. The result  on CityPersons further demonstrate the effectiveness of our approach on crowded scenes.

\section{Conclusion}
In this paper, to better handle detection in crowded scenes, we propose two new techniques, DGA for training and DG-NMS for inference, based on predicted object density maps. DGA
diminishes the negative effects of ambiguous anchors caused by overlapping bounding boxes, by jointly generating  anchor assignments and anchor reweighing via a globally OT plan matrix estimated with a density map and the proposed overlap-aware transport cost.  
The density map also provides an indicator of local crowdness and helps to adjust the NMS threshold adaptively during inference.  
We conduct extensive experiments on the heavily and moderately crowded datasets with various detector architectures, which demonstrate the effectiveness and robust of our approach to crowdness.

\bibliographystyle{IEEEtran}
\bibliography{main}

\end{document}